\pdfoutput=1
\PassOptionsToPackage{unicode}{hyperref}
\PassOptionsToPackage{hyphens}{url}
\PassOptionsToPackage{dvipsnames,svgnames,x11names}{xcolor}
\documentclass[
  11pt]{article}
\usepackage{amsmath,amssymb}
\usepackage{iftex}
\ifPDFTeX
  \usepackage[T1]{fontenc}
  \usepackage[utf8]{inputenc}
  \usepackage{textcomp} 
\else 
  \usepackage{unicode-math} 
  \defaultfontfeatures{Scale=MatchLowercase}
  \defaultfontfeatures[\rmfamily]{Ligatures=TeX,Scale=1}
\fi
\usepackage{lmodern}
\ifPDFTeX\else
\fi
\IfFileExists{upquote.sty}{\usepackage{upquote}}{}
\IfFileExists{microtype.sty}{
  \usepackage[]{microtype}
  \UseMicrotypeSet[protrusion]{basicmath} 
}{}
\makeatletter
\@ifundefined{KOMAClassName}{
  \IfFileExists{parskip.sty}{%
    \usepackage{parskip}
  }{
    \setlength{\parindent}{0pt}
    \setlength{\parskip}{6pt plus 2pt minus 1pt}}
}{
  \KOMAoptions{parskip=half}}
\makeatother
\usepackage{xcolor}
\usepackage{color}
\usepackage{fancyvrb}

\DefineVerbatimEnvironment{Highlighting}{Verbatim}{commandchars=\\\{\}}
\newenvironment{Shaded}{}{}

\newcommand{\BuiltInTok}[1]{\textcolor[rgb]{0.00,0.50,0.00}{#1}}

\newcommand{\FunctionTok}[1]{\textcolor[rgb]{0.02,0.16,0.49}{#1}}

\newcommand{\NormalTok}[1]{#1}

\usepackage{longtable,booktabs,array}
\usepackage{calc} 
\usepackage{etoolbox}
\makeatletter
\patchcmd\longtable{\par}{\if@noskipsec\mbox{}\fi\par}{}{}
\makeatother
\IfFileExists{footnotehyper.sty}{\usepackage{footnotehyper}}{\usepackage{footnote}}
\makesavenoteenv{longtable}
\usepackage{graphicx}
\makeatletter
\def\maxwidth{\ifdim\Gin@nat@width>\linewidth\linewidth\else\Gin@nat@width\fi}
\def\maxheight{\ifdim\Gin@nat@height>\textheight\textheight\else\Gin@nat@height\fi}
\makeatother
\setkeys{Gin}{width=\maxwidth,height=\maxheight,keepaspectratio}
\makeatletter
\def\fps@figure{htbp}
\makeatother
\ifLuaTeX
\usepackage[bidi=basic]{babel}
\else
\usepackage[bidi=default]{babel}
\fi
\babelprovide[main,import]{american}

\def\languageshorthands#1{}
\usepackage[T1]{fontenc}
\usepackage[utf8]{inputenc}
\usepackage{lmodern}
\usepackage{microtype}
\usepackage{booktabs,longtable,array,ragged2e,graphicx,float,caption,xcolor,enumitem}
\usepackage{geometry}
\setlist{nosep,leftmargin=*}
\AtBeginEnvironment{longtable}{\small}
\usepackage{newunicodechar}
\newunicodechar{δ}{\ensuremath{\delta}}
\newunicodechar{ρ}{\ensuremath{\rho}}
\newunicodechar{×}{\ensuremath{\times}}
\newunicodechar{→}{\ensuremath{\rightarrow}}
\newunicodechar{−}{\ensuremath{-}}
\ifLuaTeX
  \usepackage{selnolig}  
\fi
\usepackage{bookmark}
\IfFileExists{xurl.sty}{\usepackage{xurl}}{} 
\hypersetup{
  pdftitle={Passing Coarse Marginal Checks Can Be Cheap: Persona Mixtures and Imprecise Treatment-Response Estimates in an LLM Persona Panel},
  pdfauthor={Yohei Nakajima},
  pdflang={en-US},
  colorlinks=true,
  linkcolor={blue},
  filecolor={Maroon},
  citecolor={Blue},
  urlcolor={blue},
  pdfcreator={LaTeX via pandoc}}

\title{Passing Coarse Marginal Checks Can Be Cheap: Persona Mixtures and
Imprecise Treatment-Response Estimates in an LLM Persona Panel}
\author{Yohei Nakajima}
\date{Untapped Capital - July 2026}

\begin{document}
\maketitle
\begin{abstract}
Large language models are increasingly used as synthetic research
participants and are often validated by whether their marginal responses
resemble human data. We study a fixed panel of sixteen lightweight
persona-conditioned GPT-4.1 configurations in repeated strategic games.
The panel met preregistered broad-reference condition-mean criteria in
three of four repeated-game cells; the sole miss was 0.011 below the
lower reference bound. Variation was strongly prompt-indexed, but its
share depended on uncertainty assumptions: fixed-panel
symmetric-Dirichlet sensitivities produced median between-prompt shares
of 63\%-71\% under Jeffreys alpha=0.5 and 47\%-53\% under alpha=1, while
finite-opportunity plug-in estimates were 85\%-96\%. Aggregate
continuation-probability contrasts were +0.083 and +0.078, with
conservative simultaneous 95\% intervals {[}-0.171, +0.330{]} and
{[}-0.181, +0.330{]}. The treatment jointly changed the continuation
process and its textual representation. A separate wording-and-position
operation shifted cooperation from 0/40 to 37/40 in the bare
configuration, and a label conflict also revealed representation
control. The original persona-level p13 result was not prospectively
family-controlled, while a post-adjudication exact gate was structurally
underpowered; p13 is therefore a replication target rather than a
finding. External review exposed family-error, dependence, construct,
and boundary-uncertainty defects, and zero-call reanalysis changed the
interpretation without rewriting the historical record. The registered
marginal criteria could be passed without precisely estimating the
treatment-response object. A public capsule verifies 4,916 confirmatory
Phase 3-5 runs with no live model calls. The results concern one fixed
model-prompt panel and do not establish human substitutability.
\end{abstract}

\subsection{1. Introduction}\label{introduction}

Human behavioral experiments are slow and expensive; LLM calls are fast
and nearly free. Here, ``cheap'' denotes \textbf{evidentiary economy},
not merely low API cost: a validation rule can be satisfied without
requiring precise evidence about the response object it may later be
taken to support. A growing literature reports that suitably conditioned
LLMs produce data resembling human data---``algorithmic fidelity''
{[}Argyle et al.~2023{]}, ``homo silicus'' {[}Horton 2023{]}, behavior
``statistically indistinguishable from a random human'' {[}Mei et
al.~2024{]}---and a formal framework for evaluating \emph{statistical
realism} now exists {[}Xie et al.~2026{]}. Much of this evidence
validates marginals: means, distributions, and aggregate replication.

Recent work shows that descriptive realism and causal fidelity can
diverge at scale {[}Li \& Ji 2026{]}, that treating LLM outcomes as
surrogates for human outcomes requires assumptions that marginal
equivalence does not supply {[}Persson et al.~2026{]}, and that
intervention prompts can shift a model's implied latent user even when
explicit persona text is fixed {[}Lin et al.~2026{]}. Li and Ji
additionally trace effect errors to intervention logic, outcome
structure, and excessive attitude--behavior coupling, so the present
paper does not claim mechanism-level explanation in general. Its
narrower contribution is a specific fixed-panel composition mechanism in
incentive-bearing strategic interaction, coupled to minimal
representation interventions and an auditable correction record. We
decompose the explicit persona panel into between-prompt dispersion,
within-prompt variation, and prompt-indexed treatment response. The
panel passes coarse marginal checks while producing small but
imprecisely estimated continuation-probability contrasts, and much of
the recorded variation lies between prompt configurations rather than
within them. Sealed templates and matched procedures control explicit
assignment, environment randomization, and execution. They do not
establish latent-person invariance; Lin-style user drift and the
observed composition pattern can coexist.

\textbf{Contributions.} First, we provide a registered
strategic-interaction example in which a fixed persona panel passes
coarse condition-level and dispersion checks while aggregate
continuation-probability point differences remain imprecisely estimated;
the design does not establish equivalence or a narrow response bound
(§4.1). Second, we quantify the associated composition problem with
three explicitly different uncertainty views and show that the stronger
``dominant between-prompt share'' reading is prior-sensitive, while
representation experiments reveal large wording and label-linked shifts
(§4.1--4.3). Third, we demonstrate an auditable reliability
protocol---and its limits---through prospective registration, external
chronology, mechanical adjudication, complete zero-call replay of the
confirmatory record, and public correction of family-error and
construct-validity defects (§4.4--4.5).

\subsection{2. Related work}\label{related-work}

\textbf{Positioning relative to prior work.} Li and Ji {[}2026{]}
establish across three model families, eleven interventions, and 59,508
participants that descriptive fit and intervention-effect accuracy can
diverge, that prompt refinements improving realism do not reliably
improve effect accuracy, and that errors vary with intervention logic,
outcome structure, and attitude--behavior coupling. Persson,
Schultzberg, and Ankargren {[}2026{]} formalize when LLM outcomes can
serve as causal surrogates and why novel interventions still require
human evidence. Lin et al.~{[}2026{]} show that interventions can change
the implicit simulated population even when explicit personas are fixed.
Statistical-realism, persona-collapse, and state-versus-trait work
further show that persona-conditioned populations can compress or
misallocate heterogeneity {[}Xie et al.~2026; Harry et al.~2026; Xiao et
al.~2026{]}. Our differentiation is therefore not the broad divergence
or the existence of mechanisms. It is a registered decomposition of one
common lightweight construction in strategic interaction: the same fixed
explicit prompt panel is evaluated for marginal fit,
finite-opportunity-corrected between/within composition, representation
sensitivity, and response to a represented continuation-probability
treatment, with exact prompt provenance and public inferential
correction. The observed composition pattern is complementary to, not
exclusive of, latent-user drift.

\textbf{LLM strategic behavior.} Akata et al.~{[}2025{]} characterize
repeated-game play modulated by prompts; Pal et al.~{[}2026{]} elicit
strategies from five models while varying continuation probability,
payoffs, horizon knowledge, and framing; counterfactual-reasoning
evaluations alter labels and payoff structures {[}Georgousis et
al.~2026{]}; and ``strategic robustness'' has been defined as
payoff-preserving invariance across narratives {[}Mousavi Davoudi et
al.~2026{]}. These works establish that neither repeated games nor
prompt/payoff perturbations are new. Our distinct combination is the
fixed persona panel, the explicit between/within/response decomposition,
exact prompt provenance, prospective registration of confirmatory
claims, and mechanical adjudication followed by public inferential
correction.

\textbf{Strong positive evidence, and a different estimand.} Ashokkumar,
Hewitt, Ghezae, and Willer {[}2026{]} use study descriptions to forecast
469 effects from 70 preregistered, nationally representative survey
experiments and find strong correlations with realized effects,
alongside systematic effect-size overestimation and weaker performance
in a megastudy archive. That is important contrary evidence against any
blanket pessimism about LLMs in experimental science. It is also a
forecasting task over studies rather than subject-level simulation of a
response surface. Strong effect forecasting is compatible with the
fixed-panel composition failure studied here and reinforces the
prescription to validate a simulator on the exact response object for
which it will be used.

\textbf{Synthetic participants and personas.} Bisbee et al.~{[}2024{]}
find plausible survey averages alongside compressed variance, distorted
coefficients, and temporal drift; Boelaert et al.~{[}2025{]} report
excess homogeneity; Anthis et al.~{[}2025{]} catalog diversity and
generalization challenges; Hullman et al.~{[}2026{]} propose statistical
calibration for confirmatory use; and Park et al.~{[}2024{]} show that
rich interview conditioning can substantially outperform lightweight
demographic/persona descriptions. Format sensitivity {[}Sclar et
al.~2024{]}, role-play framing {[}Shanahan et al.~2023{]}, persona
collapse {[}Xiao et al.~2026{]}, state blindness {[}Harry et
al.~2026{]}, and reviews of persona-experiment transparency {[}Batzner
et al.~2025{]} all caution against treating a persona string as a stable
human analogue. RLHF-related diversity reduction is a possible mechanism
for concentrated policies, not a mechanism identified by this design. A
fuller literature map and differentiation table are included in the
public research record described in §7.

\subsection{3. Instrument and inferential
units}\label{instrument-and-inferential-units}

The primary deployment is gpt-4.1 with 16-token outputs and a fixed
minimal behavioral-subject prompt containing no game-theory vocabulary
or reasoning scaffold. Temperature was 0.7 except in the registered
Phase 5 sweep at 1.0 and 1.3. On the primary OpenAI-compatible path,
\texttt{temperature} and \texttt{max\_tokens=16} were explicitly
supplied; the assembled prompt set \texttt{top\_p=1.0}, which the
adapter intentionally omitted from the wire at 1.0, while
\texttt{presence\_penalty}, \texttt{frequency\_penalty}, and
\texttt{logit\_bias} were not supplied and therefore inherited provider
defaults. No tools or native structured output were used.

Before Phase 5 data existed, the sixteen persona sentences were
generated with a seeded \texttt{mulberry32} registry (seed 20260728) as
the full cross of two age bands and three binary trait dimensions:
agreeable/competitive, patient/impulsive, and risk-averse/risk-seeking.
The preregistered leaning label was assigned from those generated
traits---cooperative-leaning iff at least two of agreeable, patient, and
risk-averse---yielding eight prompts in each leaning stratum by
construction; no behavior was used to assign it. Names, ages, and
occupations remained uncontrolled semantic components of the complete
sealed sentences. Phase 5 prepends one such sentence to byte-identical
task text. The cross-vendor Gemini tier is descriptive; the original
Claude Haiku candidate failed a registered entry gate and was replaced
under an archived amendment. Environment randomness is seeded;
provider-side generation is not claimed to be seeded. Every request,
rendered prompt, completion, decoding configuration, round, and
provenance record is archived.

\subsubsection{3.1 Sequential architecture and registration
status}\label{sequential-architecture-and-registration-status}

Phases 1--2 developed and repaired the instrument: Phase 1 established
the historical prototype, and Phase 2 added mechanical re-adjudication
and enforcement after that prototype exposed analyst discretion. Neither
is treated as prospective confirmation.

\begin{longtable}[]{@{}
  >{\raggedright\arraybackslash}p{(\columnwidth - 6\tabcolsep) * \real{0.2500}}
  >{\raggedright\arraybackslash}p{(\columnwidth - 6\tabcolsep) * \real{0.2500}}
  >{\raggedright\arraybackslash}p{(\columnwidth - 6\tabcolsep) * \real{0.2500}}
  >{\raggedright\arraybackslash}p{(\columnwidth - 6\tabcolsep) * \real{0.2500}}@{}}
\toprule\noalign{}
\begin{minipage}[b]{\linewidth}\raggedright
Confirmatory phase
\end{minipage} & \begin{minipage}[b]{\linewidth}\raggedright
Primary role
\end{minipage} & \begin{minipage}[b]{\linewidth}\raggedright
Independence unit used here
\end{minipage} & \begin{minipage}[b]{\linewidth}\raggedright
Registration status
\end{minipage} \\
\midrule\noalign{}
\endhead
\bottomrule\noalign{}
\endlastfoot
Phase 3 & Bare GPT-4.1 repeated PD, framing, and RPS & complete episode
& claims registered before Phase 3 data \\
Phase 4 & Representation robustness, X1/X2 wording, counterfactual
labels/payoffs, continuation assays, adversaries, and sentinels &
complete episode & each block registered before its own data; X1 was
sequentially registered after Phase 3 but before any X1 data \\
Phase 5 & Sixteen sealed persona prompts crossed with Phase 4
instruments; descriptive Gemini tier & complete persona prompt, with
episodes nested beneath it & confirmatory predicates registered before
Phase 5 data; later sensitivities are explicitly post-adjudication \\
\end{longtable}

The paper's main empirical decomposition is Phase 5, interpreted using
representation results from Phases 3--4.

The full event store contains 5,505 completed runs, 54,276 round events,
108,552 seat-round decisions, and 36,251 archived provider-request
events. The public confirmatory replay contract verifies 4,916 Phase
3--5 runs: 320 registered Phase 3/X1 LLM runs, 20 deterministic Phase 3
baselines, 2,864 Phase 4 runs, and 1,712 Phase 5 runs; three additional
completed legacy entry/diagnostic runs are also replayed but are not
counted as confirmatory. A separate transactional ledger records 30,530
Phase 4--5 calls, 13,141,675 input tokens, and 45,247 output tokens. The
low output-token average (about 1.5 per call) is expected because valid
actions were ordinarily one-token completions even though
\texttt{max\_tokens=16}; the ledger excludes earlier phases and must not
be conflated with the full event-store request count. Machine-readable
count definitions are included in the public record (§7).

Confirmatory claims were registered before the data that adjudicated
them and were mechanically evaluated in a fixed vocabulary. The
historical two-sided interiority rule used Clopper--Pearson bounds on
seat-level round-one trials. Because two seats share an episode, the
submission analysis treats the complete episode as the independence
unit. For prompt \(i\), condition \(d\), and complete episode \(e\), let
\(C_{ide}\in\{0,1,2\}\) be the number of cooperating seats,
\(A_{ide}=\mathbf 1(C_{ide}\ge1)\), and
\(B_{ide}=\mathbf 1(C_{ide}=2)\). The episode mean is then
\(Y_{ide}=(A_{ide}+B_{ide})/2\in\{0,0.5,1\}\). Simultaneous
Clopper--Pearson intervals are constructed for the two binary components
and projected onto \(E[Y]\). For six episodes,
\(\hat p_i(d)=\{\bar A_i(d)+\bar B_i(d)\}/2\). For each aggregate
contrast, sixteen equally weighted prompt cells contribute 96 episodes
per condition; pooled component counts estimate
\(\bar p(d)=16^{-1}\sum_i p_i(d)\). Four component-condition intervals
split the total error rate by Bonferroni. If \([L_d,U_d]\) is the
projected condition-mean interval, the contrast interval is

\[
[L_{.90}-U_{.10},\;U_{.90}-L_{.10}].
\]

Under the pooled independent-binomial component model this union-bound
construction has at least 95\% simultaneous coverage. With independent
heterogeneous prompt probabilities, the component totals are
Poisson-binomial and no more dispersed than the equal-probability
binomial at the same mean, making the binomial projection conservative
for that working model. Its role is a conservative small-sample
projection, not a claim that the sixteen prompt propensities are
homogeneous. It does not assume seat independence and does not collapse
to zero uncertainty when all recorded episodes agree. A fixed-panel
symmetric-Dirichlet sensitivity assigns each prompt-cell outcome
distribution a prior over \(\{0,0.5,1\}\) and projects
\(E[Y]=0.5q_{0.5}+q_1\). The primary sensitivity uses
Dirichlet(0.5,0.5,0.5); post-adjudication prior sensitivity also reports
symmetric \(\alpha=0.25\) and \(\alpha=1\). The percentile cluster
bootstrap is retained as a separate sensitivity.

The hierarchy is deployment → explicit persona prompt → condition →
episode → seat → round → provider request. Phase 5's confirmatory unit
is the complete persona sentence; name, age, occupation, and traits are
bundled semantic treatments. Registered claims attach to the conditional
finite-panel estimand for these sixteen prompts. Claims about a wider
persona generator are exploratory at \(n=16\). Pairing the same explicit
prompt across conditions identifies a prompt-indexed contrast, not
necessarily a stable latent person's treatment effect.

\textbf{Protocol glossary.} \texttt{S2-absent} and \texttt{S2-present}
are the registered repeated-game wording families.
\textbf{Switch-bearing} means the span whose adjacent substitution
produced the largest preregistered ladder gap and subsequently passed
held-out confirmation; S2-present contains that
replacement-and-reposition operation, while S2-absent contains the
original sentence. \texttt{P3-A3} is the Phase 3 registered
broad-reference cooperation claim, with band {[}0.36, 0.63{]}.
\texttt{P5-1a}, historically called the \textbf{corner-mixture
predicate}, is the registered support condition that fires when the
interior fraction in the exact-bare-twin restricted set is below 0.10
under the frozen seat-level rule; it is not a general theorem about
mixture structure. \texttt{P5-1b} is the registered between-persona
dispersion comparison. \texttt{P5-2} pools registered conflict cells and
classifies whether choices follow task text or persona-conditioned
direction. \texttt{P5-3(a)} asks whether any persona × wording pair has
both continuation-probability cells interior and a positive slope lower
bound; \texttt{P5-3(b)} asks whether each persona lane rejects the bare
configuration's dominated swap-cell option at a registered minimum rate.
Historical verdict labels remain visible even where post-adjudication
analyses change their scientific interpretation.

\textbf{Phase 5 condition matrix.} The 96 Tier-A persona--condition
units are the full cross of sixteen prompts with six conditions:

\begin{longtable}[]{@{}
  >{\raggedright\arraybackslash}p{(\columnwidth - 4\tabcolsep) * \real{0.3333}}
  >{\raggedright\arraybackslash}p{(\columnwidth - 4\tabcolsep) * \real{0.3333}}
  >{\raggedright\arraybackslash}p{(\columnwidth - 4\tabcolsep) * \real{0.3333}}@{}}
\toprule\noalign{}
\begin{minipage}[b]{\linewidth}\raggedright
Code
\end{minipage} & \begin{minipage}[b]{\linewidth}\raggedright
Condition
\end{minipage} & \begin{minipage}[b]{\linewidth}\raggedright
Role in the paper
\end{minipage} \\
\midrule\noalign{}
\endhead
\bottomrule\noalign{}
\endlastfoot
\texttt{rep-d10-s2a} & repeated PD, δ=.10, S2 absent & repeated-game
level, variance, and response \\
\texttt{rep-d10-s2p} & repeated PD, δ=.10, S2 present & repeated-game
level, variance, and response \\
\texttt{rep-d90-s2a} & repeated PD, δ=.90, S2 absent & repeated-game
level, variance, and response \\
\texttt{rep-d90-s2p} & repeated PD, δ=.90, S2 present & repeated-game
level, variance, and response \\
\texttt{os-swap} & one-shot canonical-payoff label swap &
semantic-label/payoff conflict \\
\texttt{os-community} & one-shot Community framing & near-interior
framing anchor \\
\end{longtable}

The registered P5-1a denominator was restricted to persona cells whose
exact recorded bare twin failed the same interiority gate. An
outcome-blind exact-twin completion fixed that set as
\texttt{rep-d90-s2a} and \texttt{os-swap}: sixteen personas in each
condition, hence 32 units. The Community twin passed the bare gate; the
other three repeated-game cells lacked exact bare twins and entered only
the unrestricted 96-cell secondary.

The machinery's boundary is explicit:

\begin{quote}
\textbf{The pipeline can enforce a registered predicate exactly; it
cannot guarantee that the predicate represents a valid estimand, test
family, or construct.}
\end{quote}

\subsection{4. Results}\label{results}

\subsubsection{4.1 Prompt composition is substantial; treatment response
remains loosely
bounded}\label{prompt-composition-is-substantial-treatment-response-remains-loosely-bounded}

Across the four repeated-game cells, the fixed-panel Jeffreys
\(\alpha=0.5\) sensitivity yields median between-prompt shares of
63.1\%, 70.5\%, 66.1\%, and 66.6\%, with 95\% intervals {[}49.4\%,
74.5\%{]}, {[}57.3\%, 81.3\%{]}, {[}52.7\%, 77.1\%{]}, and {[}52.7\%,
77.7\%{]}. The stronger claim that between-prompt variation exceeds
one-half is not prior-robust. In a zero-call symmetric-prior sweep,
\(\alpha=0.25\) produces medians of 74.8\%--82.5\% and
\(P(B/(B+W)>0.5)>0.999\) in every cell; Jeffreys \(\alpha=0.5\) produces
medians of 63.1\%--70.5\% with probabilities 0.970--0.998; and
\(\alpha=1\) produces medians of 47.1\%--53.5\% with probabilities
0.325--0.700. Thus the data support substantial prompt-indexed
composition across the sweep, while ``dominant between-prompt share'' is
prior-dependent.

For comparison, finite-opportunity plug-in shares are 85.5\%, 96.1\%,
88.8\%, and 90.2\%, with conditional episode-bootstrap 95\% intervals
{[}82.0\%, 93.8\%{]}, {[}94.6\%, 98.9\%{]}, {[}86.7\%, 94.6\%{]}, and
{[}87.9\%, 95.5\%{]}. The plug-in view conditions strongly on recorded
boundary concentration and tends upward; symmetric-Dirichlet posteriors
shrink six-episode corner cells toward the interior and tend downward.
They are bracketing descriptions of different uncertainty assumptions,
not interchangeable estimators. Figure~1 compares the two composition
views across all four repeated-game cells.

\includegraphics[width=0.95\textwidth,height=\textheight]{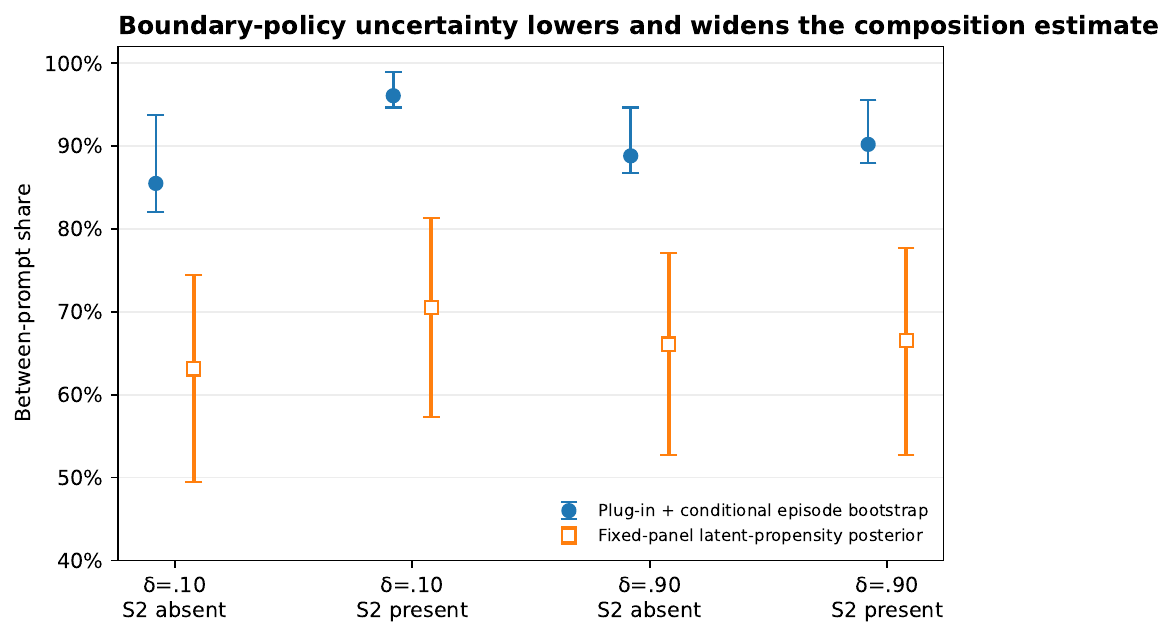}

\emph{Figure~1. Between-prompt share of episode-level variation. The
Jeffreys fixed-panel posterior propagates uncertainty in six-episode
prompt cells; plug-in/conditional-bootstrap estimates describe the
archived concentration more literally. A symmetric-prior sweep reported
in text shows that the stronger ``share above one-half'' interpretation
is prior-dependent.}

Across the represented continuation-probability treatment, the observed
fixed-panel point differences are +0.083 for S2-absent wording and
+0.078 for S2-present wording. Conservative exact simultaneous 95\%
intervals are {[}−0.171, +0.330{]} and {[}−0.181, +0.330{]}. Their
breadth jointly reflects the registered six-episode-per-cell design and
an exact projection that retains non-zero uncertainty at empirical
corners. The treatment changes both the continuation process and the
text used to communicate it; round-one actions identify response under a
specified representation, not a semantically neutral economic parameter.
The point estimates are small on the unit scale, but the data do not
establish equivalence, a zero response, or a narrow upper bound.

The empirical point is not that six repeated episodes should have
produced high precision. It is that the registered marginal criteria did
not require the treatment-response object to be estimated precisely, yet
their passage could be read as validation; the intervals quantify what
that validation design left unidentified. For the finite archived panel,
+0.083 and +0.078 are exact descriptive arithmetic. A design-effect
heuristic using six episodes per prompt and the plug-in between-share
range 0.855--0.961 gives roughly 16.5--18.2 episode equivalents per
condition, close to the sixteen prompt units. This is not the degrees of
freedom of the exact procedure, but it identifies prompt-family size as
the operative precision constraint. Figures~2--3 show the prompt-indexed contrasts
and fixed-panel condition means.

\includegraphics[width=0.95\textwidth,height=\textheight]{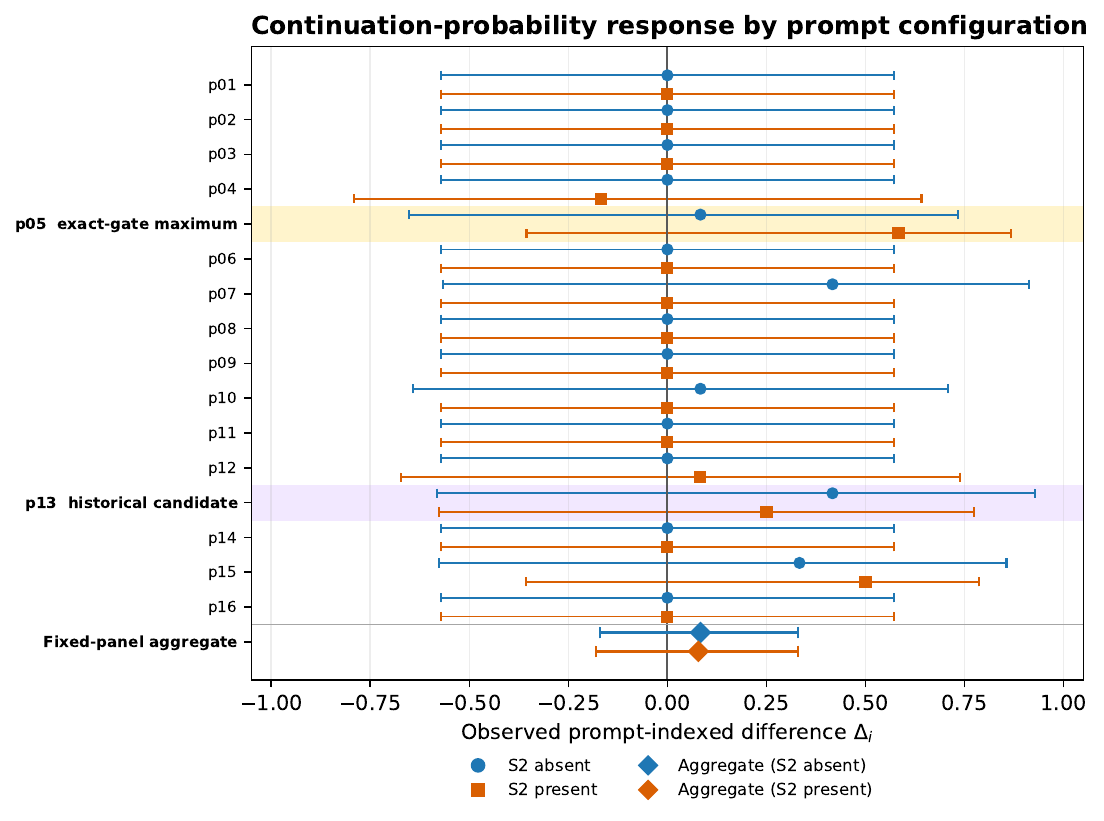}

\emph{Figure~2. Prompt-indexed differences in round-one cooperation,
\(\Delta_i=\hat p_i(\delta=.90)-\hat p_i(\delta=.10)\), for both wording
families. Bars are conservative exact simultaneous 95\% intervals with
complete episodes as the unit; observed corners retain non-zero
uncertainty. The p05 row is flagged as the largest candidate eligible
under the conservative exact family gate, and p13 as the historical
candidate selected by the original rule.}

\includegraphics[width=0.95\textwidth,height=\textheight]{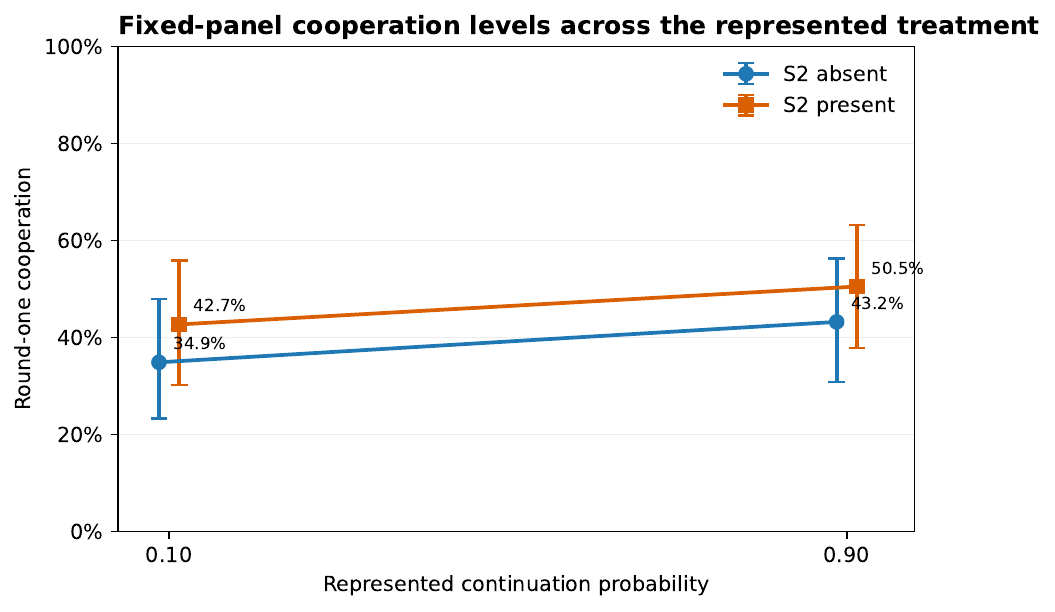}

\emph{Figure~3. Fixed-panel cooperation by represented
continuation-probability condition. Error bars are conservative exact
condition intervals; lines connect conditions for orientation only.}

The broad-reference cooperation band was {[}0.36, 0.63{]}. Repeated-game
pool means were 0.349, 0.427, 0.432, and 0.505; only the S2-absent
\(\delta=.10\) cell fell outside, by 0.011 below the lower boundary. The
preregistered leaning rule divides the panel into eight
cooperative-leaning and eight defect-leaning complete prompts.
Descriptive gaps range from 0.510 to 0.719 across the repeated-game
conditions and are fixed-panel prompt-bundle contrasts, not causal trait
effects:

\begin{longtable}[]{@{}
  >{\raggedright\arraybackslash}p{(\columnwidth - 8\tabcolsep) * \real{0.1579}}
  >{\raggedleft\arraybackslash}p{(\columnwidth - 8\tabcolsep) * \real{0.2105}}
  >{\raggedleft\arraybackslash}p{(\columnwidth - 8\tabcolsep) * \real{0.2105}}
  >{\raggedleft\arraybackslash}p{(\columnwidth - 8\tabcolsep) * \real{0.2105}}
  >{\raggedleft\arraybackslash}p{(\columnwidth - 8\tabcolsep) * \real{0.2105}}@{}}
\toprule\noalign{}
\begin{minipage}[b]{\linewidth}\raggedright
condition
\end{minipage} & \begin{minipage}[b]{\linewidth}\raggedleft
cooperative-leaning mean
\end{minipage} & \begin{minipage}[b]{\linewidth}\raggedleft
defect-leaning mean
\end{minipage} & \begin{minipage}[b]{\linewidth}\raggedleft
difference
\end{minipage} & \begin{minipage}[b]{\linewidth}\raggedleft
prompts per stratum
\end{minipage} \\
\midrule\noalign{}
\endhead
\bottomrule\noalign{}
\endlastfoot
\texttt{rep-d10-s2a} & 0.615 & 0.083 & +0.531 & 8 \\
\texttt{rep-d10-s2p} & 0.760 & 0.094 & +0.667 & 8 \\
\texttt{rep-d90-s2a} & 0.688 & 0.177 & +0.510 & 8 \\
\texttt{rep-d90-s2p} & 0.865 & 0.146 & +0.719 & 8 \\
\texttt{os-community} & 0.688 & 0.019 & +0.669 & 8 \\
\end{longtable}

P5-1b used protocol-nonmatched human SD references mechanically implied
from Dal Bó and Fréchette's {[}2011{]} R=40 strategy-frequency
estimates: 0.4122 for their \(\delta=.50\) panel and 0.3116 for
\(\delta=.75\). The frozen ratio \(\rho=.75\) gave thresholds 0.3092 and
0.2337. It was a preregistered heuristic tolerance, not an equivalence
margin. Finite-opportunity-corrected plug-in SDs were 0.4182, 0.4784,
0.4408, and 0.4323. A conditional episode bootstrap produced
corrected-SD intervals {[}0.4122, 0.4391{]}, {[}0.4696, 0.4916{]},
{[}0.4279, 0.4654{]}, and {[}0.4269, 0.4496{]}, but unanimous cells
remain point masses in that bootstrap. The first lower bound was
independently reproduced at full precision; its displayed equality to
the 0.4122 human reference is a rounding coincidence. A two-stage
prompt+episode bootstrap changes the estimand toward a hypothetical
persona generator and yields wider SD intervals {[}0.2724, 0.4879{]},
{[}0.3696, 0.5123{]}, {[}0.3457, 0.4890{]}, and {[}0.3345, 0.4847{]}.
P5-1b is therefore retained as a permissive historical checkpoint, not
evidence of human-variance equivalence.

Across all six Phase 5 conditions, the historical seat-level rule
classifies 14/96 persona-condition cells interior, the exact episode
projection 11/96, and the Jeffreys sensitivity 19/96. In the registered
32-unit set, the counts are 3/32, 2/32, and 5/32. At \(n=6\) with a
three-valued outcome, modest differences in interval width move cells
across the threshold; continuous uncertainty and variance components are
more informative than the binary census.

Descriptively, the composition pattern did not transfer to the
endpoint-nonstationary Gemini tier: 9/24 Gemini cells met the historical
interiority rule versus 14/96 in the primary panel, and several
representation effects reversed direction. This is consistent with
deployment-specific composition; the tier is not a formal replication.
Human references remain protocol-nonmatched because continuation
probabilities, payoffs, incentives, assignment, and experience differ
from the present design; no matched magnitude or human-equivalence claim
is made.

\subsubsection{4.2 Representation interventions expose prompt-surface
control}\label{representation-interventions-expose-prompt-surface-control}

For the bare configuration, round-one cooperation was 0.000 at every
registered continuation probability. X2 decomposed the v1 and v2a prompt
bundles into six sentence/block spans and constructed forward and
reverse ladders by replacing one complete span at a time. The selected
S2 operation replaced and repositioned ``After every round there is a
\{deltaPct\}\% chance the session continues with another round'' with
``At the end of each round there is a \{deltaPct\}\% chance that the
session goes on for one more round.'' Screening used ten episodes per
rung; the selected minimal pair was confirmed at temperature 0.7 on 20
fresh episodes per side (seeds 2953--2972), moving cooperation from 0/40
to 37/40. Because wording and position were one atomic operation, the
design does not separate them.

The effect was specific to the representation of repeated interaction,
not a generic main effect of wording. In the 640-episode one-shot D1
battery, GPT-4.1's registered wording main effect was +0.0063
(SE 0.0210; Holm-adjusted $p=1.00$), and none of the registered wording
interactions was supported. Conversely, the D1 presentation selected for
the repeated assay had a one-shot mean of 0.100; when embedded in the
repeated-game protocol, round-one cooperation rose to 0.750 at
$\delta=.10$ and 1.000 at $\delta=.90$. The Community wrapper produced
1.000 in both conditions. These repeated cells were correctly classified
as ceiling-confounded, so they do not identify a continuation-probability
slope. They do show that announcing and implementing repeated interaction
can be a much larger treatment than ordinary one-shot wording variation.

In the label-swap cell, canonical payoffs were held fixed while
``Cooperate'' and ``Defect'' were attached to opposite strategic roles.
The bare configuration chose the cooperation-worded option 0/40 times
and instead took the payoff-dominated role whenever it carried the word
``Defect.'' This shows that the displayed label or a label-linked
learned prior can override payoff dominance in this registered cell. It
does not identify intrinsic lexical valence: a learned game-theoretic
association such as ``Defect = equilibrium/dominant action in Prisoner's
Dilemma'' is an equally plausible mechanism. A structurally equivalent
non-PD control retaining the same labels was not run.

Persona conditioning produces two observed contrasts, but they are not
factorially separable. Differences among complete persona prompts
generate the leaning gaps reported in §4.1. Adding any tested persona
string reverses the bare swap-cell choice, yet no non-semantic prefix
matched for length, punctuation, and position was run; semantic persona
content cannot be isolated from generic sequence-format disruption. In
the swap cell, label and payoff also point to the same option for
persona-conditioned configurations, leaving the reversal mechanism
ambiguous.

P5-3(b)'s 24 evaluable lanes comprise sixteen personas at T=0.7 plus
p02, p06, p11, and p15 at each of T=1.0 and T=1.3. Every lane retains a
simultaneous episode-exact lower bound above the frozen 0.20 threshold;
the minimum is 0.462. The pooled P5-2 task-consistent share is 90/704
seat decisions across 352 episodes, equivalently 45/352=0.128 on episode
means. The frozen historical adjudication used the seat-level rule. An
episode-iid Clopper--Pearson projection gives {[}0.092, 0.172{]}, but
that interval does not propagate the prompt clustering visible elsewhere
in the study. A post-adjudication stratified prompt-cluster bootstrap over
the forty registered persona × conflict-cell clusters preserves the
empirical point estimate and gives 95\% interval {[}0.071, 0.189{]}.

Independent fixed-panel Dirichlet aggregation is materially
prior-sensitive because every sparse cluster receives its own symmetric
prior. Under the Jeffreys specification, each cluster receives total prior
concentration 1.5, or 60 category-count units across forty clusters; this
is not literally equivalent to sixty pooled episodes because the clusters
have unequal sizes, but it is non-negligible relative to 352 observations
and pulls the aggregate toward 0.5. A zero-call sensitivity sweep gives:

\begin{center}
\begin{tabular}{rccc}
\toprule
symmetric $\alpha$ & posterior median & 95\% interval & $P(\theta\le .20)$ \\
\midrule
0.10 & 0.138 & {[}0.124, 0.153{]} & 1.000 \\
0.25 & 0.152 & {[}0.135, 0.171{]} & $>$0.999 \\
0.50 & 0.172 & {[}0.152, 0.195{]} & 0.991 \\
1.00 & 0.205 & {[}0.182, 0.231{]} & 0.329 \\
\bottomrule
\end{tabular}
\end{center}

The Jeffreys posterior's proximity to the registered 0.20 boundary is
therefore prior-dependent rather than an independent signal from the data;
the $\alpha=1$ posterior crosses the boundary. The prompt-cluster
bootstrap is the principal dependence-aware sensitivity. Every repeated
conflict subcell is mixed; only the swap cell is individually
persona-dominant. The historical pooled classification is therefore
mechanism-confounded and carried by the swap cell, not evidence of a
general persona-dominance mechanism. Figure~4 summarizes the two
bare-configuration representation interventions.

\includegraphics[width=0.95\textwidth,height=\textheight]{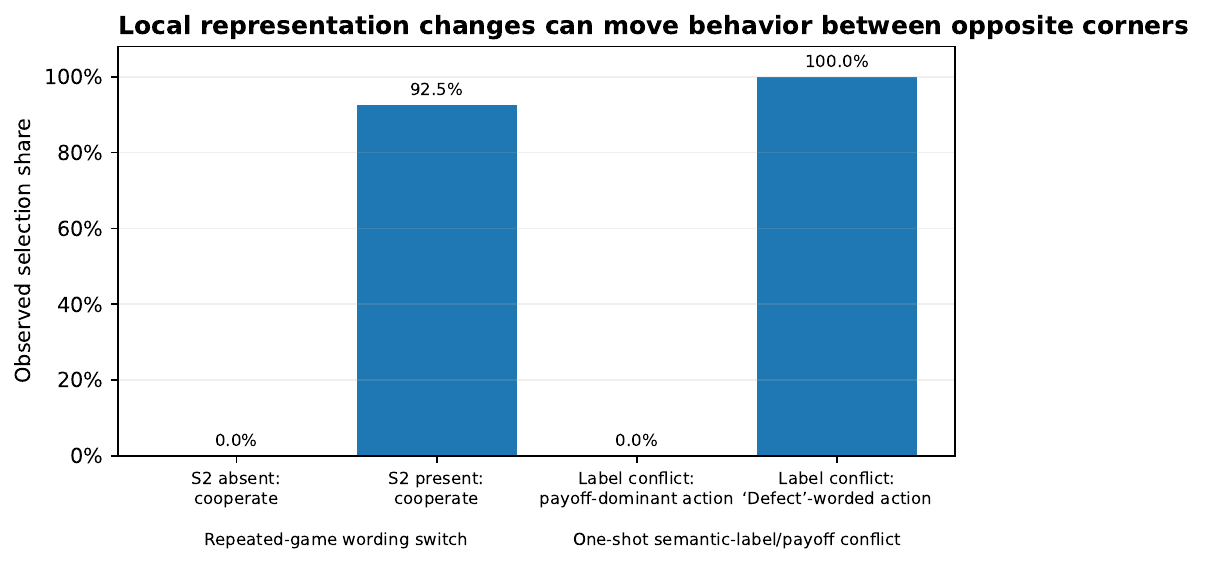}

\emph{Figure~4. Two representation interventions in the bare
configuration. The S2 wording-and-position operation moved cooperation
from 0/40 to 37/40. In the one-shot label conflict, the payoff-dominant
action was never chosen when the dominated role carried ``Defect.''
These bars report selection shares, not a common or uniquely identified
causal mechanism.}

\subsubsection{4.3 What the marginal checks cannot
identify}\label{what-the-marginal-checks-cannot-identify}

Let \(p_i(d)=E[Y\mid i,d]\), where \(i\) indexes the complete explicit
persona prompt and \(d\) the experimental condition.

\textbf{Proposition A: broad bands only partially identify the aggregate
contrast.} If a synthetic condition mean is accepted within tolerance
\(\epsilon_d\) of a reference mean in each condition, then

\[
|\Delta^S-\Delta^H|\le \epsilon_0+\epsilon_1.
\]

Equivalently, accepted bands \([\ell_0,u_0]\) and \([\ell_1,u_1]\) imply
only

\[
\Delta^S\in[\ell_1-u_0,u_1-\ell_0].
\]

Exact condition-specific mean matching would force the aggregate effect
by the identity \(\Delta=\mu_1-\mu_0\); the empirical failure lives in
the slack of coarse criteria.

\textbf{Proposition B: aggregate moments do not identify microstructure
or response coupling.} This is an application of the law of total
variance and classical Fréchet--Hoeffding/Sklar coupling results
{[}Hoeffding 1940; Sklar 1959{]} to synthetic-participant validation,
not a new probability theorem. Mean and total variance do not identify
how variation is divided between prompt configurations and repeated
draws, nor do they identify distributional shape or boundary
concentration. Even exact condition-specific distributions do not
identify the cross-condition coupling and therefore do not identify the
distribution of prompt-indexed responses \(\Delta_i=p_i(1)-p_i(0)\).
Reusing an explicit persona string supplies one prompt-indexed coupling,
but interpreting it as a stable synthetic individual's potential-outcome
contrast requires latent-person invariance, which this study does not
test.

The study therefore identifies a composition pattern in one fixed prompt
panel. It does not establish that humans have a different
microstructure, that RLHF caused the pattern, that latent-user drift is
absent, or that the same pattern occurs across persona generators.

\subsubsection{4.4 The favored persona-level result is not prospectively
confirmed; the archived family is
underpowered}\label{the-favored-persona-level-result-is-not-prospectively-confirmed-the-archived-family-is-underpowered}

Under the historical seat-level rule, persona p13 moved from 0.333
cooperation at δ=.10 to 0.750 at δ=.90 and passed a per-candidate
lower-bound test. The rule searched multiple persona × wording
candidates and fired on any pass without declared family-level error
control. External review identified that defect.

Three 200,000-permutation gate constructions are now reported; all
permutation p-values use the add-one convention,
\(\widehat p=(r+1)/(B+1)\), with exact Monte Carlo intervals. Under the
historical seat-level gate, p13 remains the maximum at +0.4167, with
familywise \(p=0.059230\), Monte Carlo 95\% interval {[}0.058194,
0.060268{]}. Under the percentile episode-cluster-bootstrap sensitivity,
p13 also remains the maximum and \(p=0.043455\), interval {[}0.042561,
0.044353{]}. The exact projection is the conservative reference because
it has finite-sample coverage for the discrete episode mean; the
percentile bootstrap is retained symmetrically but has no comparable
small-sample coverage guarantee. Under the conservative exact-episode
gate, p13 is ineligible: its low-δ lower bound falls below 0.05 and its
high-δ upper bound exceeds 0.95. Only p04/s2p and p05/s2a pass both
gates; the largest eligible slope belongs to p05/s2a (+0.0833), with
familywise \(p=0.773206\), interval {[}0.771363, 0.775039{]}.

For p13/s2a, the percentile bootstrap admitted both conditions as
interior---δ=.10: {[}0.083, 0.667{]}; δ=.90: {[}0.583,
0.917{]}---whereas the conservative exact projection rejected
both---{[}0.047, 0.800{]} and {[}0.287, 0.954{]}, respectively. Neither
recorded cell was at an exact corner. The eligibility difference
therefore arises from small-sample interval width and coverage behavior,
not from a corner interval falsely passing the gate.

The complete data-dependent gate is dynamically reapplied within every
permutation, not frozen from the observed-data mask. The implementation
precomputes 56 possible-composition gate values and performs 25,600,000
condition-gate lookup applications at B=200,000. In a deliberately
incorrect comparison that froze the observed-data mask, the maximum
statistic differed from the dynamic procedure in 718 of 5,000 null draws
(14.4\%), showing that reapplication materially changes the reference
distribution. Lookup/direct parity and the frozen-mask regression are
included in the public audit record (§7).

An exhaustive attainability audit shows that, with six episodes per
condition, 12 of the 28 possible episode-value compositions pass the
exact gate. Their sample means range from 0.333 to 0.667, but
eligibility depends on the full \({0,0.5,1}\) composition rather than on
the mean alone. Two eligible cells can therefore differ by at most
0.333. Under the archived 32-candidate null structure, the add-one
familywise tail probability at that maximum attainable slope is
0.075040, with Monte Carlo 95\% interval {[}0.073884, 0.076198{]}; no
exact-gate result in this archived family can reach 0.05. The exact
analysis is therefore a valid dependence-aware sensitivity but not a
powered disconfirmation of a p13-sized capability claim. The record
neither prospectively confirms nor decisively disconfirms p13; it
identifies a replication target whose next test must be sized
prospectively (Appendix A.3). A Phase 6 test will preregister the
candidate family, episode-level dependence unit, interiority gate,
maximum statistic, familywise decision rule, and sample size before any
data are collected.

These variants were specified and executed together only after external
review exposed the original defects. No familywise gate was registered
at the original freeze, and the repository contains no
seal-before-compute record for these sensitivities. The analyses ran in
GitHub Actions against archived databases with fixed seeds, and their
outputs were committed regardless of direction. Accordingly, the
favorable \(p=0.043455\) variant cannot create prospective confirmation,
just as the other variants cannot retroactively strengthen the frozen
claim. The historical mechanical P5-3 verdict remains visible because
clause (b) fired strongly and p13 passed the rule as frozen. Clause (a)
did not prospectively establish the capability claim because the frozen
search lacked family control; the conservative post-adjudication
procedure is too underpowered at n=6 to provide decisive evidence
against it. Figure~5 compares the three post-adjudication familywise
constructions and their different eligibility rules.

\includegraphics[width=0.95\textwidth,height=\textheight]{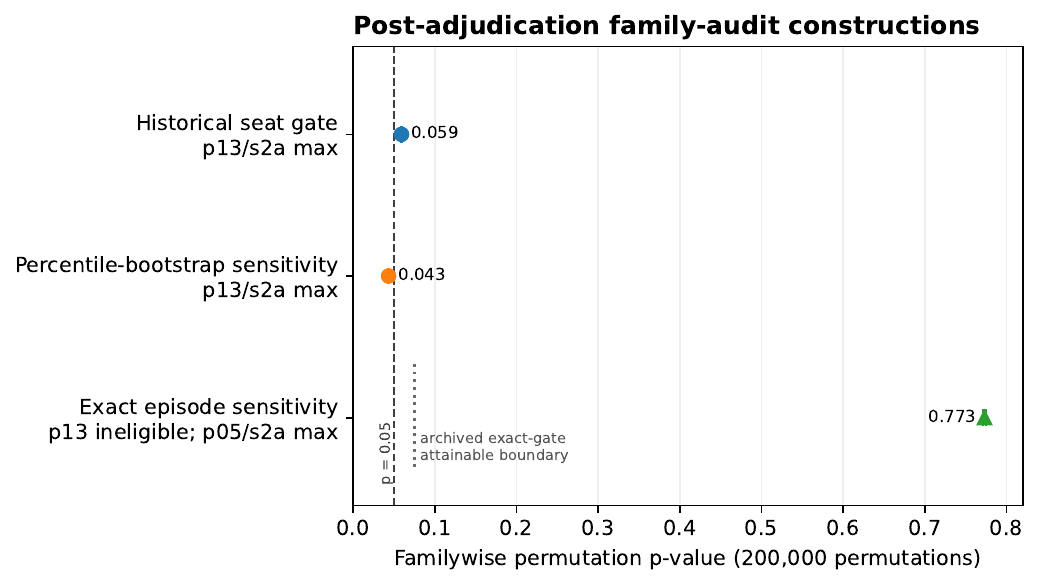}

\emph{Figure~5. Post-adjudication familywise constructions. The first
two points are p13/s2a under the historical and percentile-bootstrap
gates. Under the conservative exact-episode gate, p13 is ineligible; the
third point is the familywise result for the largest eligible candidate,
p05/s2a. The dotted line at \(p=0.075040\) marks the estimated minimum
attainable familywise p-value for the archived \(n=6\), 32-candidate
exact-gate design and applies only to that construction. None of the
procedures was registered at the original freeze; see §4.4 for the
bounded scientific status.}

\subsubsection{4.5 Auditability}\label{auditability}

Twelve registered author predictions were refuted by data and published.
Four underspecified analysis choices were resolved outcome-blind. A
two-sided gate blocked a false ceiling conclusion. Sentinels detected a
time-indexed behavioral discontinuity in an unversioned endpoint and
exposed a monitoring gap that was repaired with an attestation gate. The
public capsule now verifies all 4,916 confirmatory Phase 3--5 runs with
zero live model calls: 4,896 LLM runs replay byte-exact and 20
deterministic baselines are independently recomputed. External review
then found the family-error and dependence defects discussed above; the
archived record made both diagnosable and correctable.

The claim is bounded: the machinery is procedurally exact and
extensively auditable. Its strongest achievement is not self-validation
but preservation of enough provenance for outsiders to identify where
procedural correctness stopped short of statistical validity.

\subsection{5. Discussion}\label{discussion}

\subsubsection{5.1 Implications}\label{implications}

For synthetic-participant practice, broad marginal resemblance is a weak
validation target. Lightweight conditioning can generate plausible
aggregate levels and substantial cross-prompt dispersion while yielding
small observed treatment-response point estimates whose uncertainty
remains wide. Validation should therefore report response surfaces over
declared interventions and representation families, together with
assay-sensitivity checks, dependence-aware uncertainty, model/provider
provenance, and temporal monitoring. Statistical calibration {[}Hullman
et al.~2026{]}, causal-surrogacy assumptions {[}Persson et al.~2026{]},
and latent-drift diagnostics {[}Lin et al.~2026{]} are complementary
rather than competing safeguards.

The result can be read as a synthetic-subject analogue of the
reduced-form/structural distinction associated with the Lucas critique
{[}Lucas 1976{]}: fitting or selecting for aggregate resemblance need
not identify behavior under a changed treatment. In psychometric terms,
it is a construct-validity and assay-sensitivity problem {[}Cronbach \&
Meehl 1955; ICH E10; Temple \& Ellenberg 2000{]}. Agent-based modeling's
equifinality and pattern-oriented validation provide a related analogy
{[}Windrum et al.~2007; Grimm et al.~2005{]}. These are organizing
analogies, not claims that the study literally estimates a structural
economic model or validates a human measurement model.

The results also show why binary certification language should be used
sparingly. The historical P5-1a predicate passes under its frozen
seat-level rule and under the conservative episode-exact interval, but
fails under a reasonable Dirichlet--Jeffreys sensitivity. The underlying
continuous evidence is clearer than the thresholded label: plug-in
estimates condition on empirical boundary concentration and therefore
tend upward, whereas the Jeffreys posterior shrinks six-episode corner
cells toward the interior and therefore tends downward. These views
bracket the composition claim under opposite conditioning choices rather
than compete as interchangeable estimates. Under the Jeffreys prior, the
posterior medians remain above one-half while the lowest 95\% bound
crosses it; under the symmetric uniform prior, the medians themselves
straddle one-half. The stronger dominance interpretation is therefore
prior-dependent. P5-2 provides a second example: its historical mechanical
verdict is preserved, but the fixed-panel Bayesian proximity to the 0.20
boundary depends on the symmetric cluster prior, and the $\alpha=1$
posterior crosses that boundary.

For deployment, behavior that can be rewritten by a sentence, an action
token, or an identity prefix is safety-relevant. The label-conflict
result does not show that lexical valence always dominates incentives:
numerical payoffs move behavior in other cells, and the observed choice
may reflect semantic framing, learned game-theoretic priors, or their
interaction.

\subsubsection{5.2 The precommitted discussion, and what changed after
review}\label{the-precommitted-discussion-and-what-changed-after-review}

The Phase 5 discussion was sealed before its data existed. Excerpt (full
text in supplement; sha \texttt{1f1d7de9…e356}):

\begin{quote}
``The headline of Phase 5 is an existence result the program registered
against itself: at least one persona in the sealed sixteen passed the
two-sided assay gate and showed the registered signature of incentive
sensitivity. The author's registered prediction---that none would---is
refuted, and the refutation is the finding. \ldots{} The capability was
recoverable by content-side conditioning\ldots{} The scope of the claim
is deliberately narrow.''
\end{quote}

\begin{longtable}[]{@{}
  >{\raggedright\arraybackslash}p{(\columnwidth - 2\tabcolsep) * \real{0.5000}}
  >{\raggedright\arraybackslash}p{(\columnwidth - 2\tabcolsep) * \real{0.5000}}@{}}
\toprule\noalign{}
\begin{minipage}[b]{\linewidth}\raggedright
Precommitted interpretation
\end{minipage} & \begin{minipage}[b]{\linewidth}\raggedright
Current status after external review and zero-call reanalysis
\end{minipage} \\
\midrule\noalign{}
\endhead
\bottomrule\noalign{}
\endlastfoot
At least one persona establishes an unconfounded incentive-response
existence result & \textbf{Not prospectively established.} The frozen
search lacked family control, and the post-adjudication procedures were
method-dependent and unregistered. See §4.4 for the power-bounded
current status. \\
``No game-relevant instruction---trait words only'' & Restated
precisely: no explicit game terminology, action recommendation, or
payoff reference; traits are strategically relevant information, and
name, age, and occupation are uncontrolled semantic treatments. \\
Persona framing ``contested or beat'' task switches & Pooled choice
result survives exact episode inference, but every repeated conflict
subcell is mixed; the pooled dominance classification is entirely
carried by the word/payoff-confounded swap cell. \\
Bare corners characterize the configuration rather than the model's
capability envelope & Retained only as a future hypothesis. The p13
evidence no longer supports it; clause (b) demonstrates a robust choice
reversal but does not identify incentive sensitivity. \\
\end{longtable}

The sealed text remains unchanged because its evidentiary value lies
partly in making interpretive error visible. The correction is additive
and explicit rather than silently rewritten.

\subsection{6. Limitations}\label{limitations}

The confirmatory predicates were registered and adjudicated separately
at nominal thresholds; no study-wide alpha allocation or familywise rule
was registered across P5-1a, P5-1b, P5-2, P5-3(a), P5-3(b), and the
sequential X1 extension. They address distinct estimands and are not
interpreted as one omnibus test, but study-level false-positive exposure
is greater than under a prospectively hierarchical design. Appendix A.3
specifies the prospective correction.

The primary evidence comes from one deployment and sixteen complete
prompt bundles. Generalization to a persona generator, other models, or
human participants is not identified. The uncertainty views answer
different questions: symmetric-Dirichlet posteriors are prior-dependent,
the plug-in/conditional bootstrap conditions on recorded boundary
concentration, and the two-stage bootstrap changes the estimand by
resampling prompts. The prior sweeps in §§4.1 and 4.2 make the dependence
of the ``dominant composition'' and P5-2 boundary readings explicit.

Treatment-response intervals are wide because each prompt-cell has six
independent episodes and the exact projection retains uncertainty at
empirical corners; the binary census is correspondingly sensitive to
small-n discrete interval width. Explicit persona strings are paired
across conditions, but latent-person invariance is untested. The
continuation process was manipulated together with its textual
representation. The persona-prefix contrast lacks a format-matched
neutral control. The label conflict cannot distinguish semantic valence
from memorized game-theoretic associations because no non-PD control
retained the same labels.

The registered choice-entropy secondary is defined in Appendix A.1. Its
matched-lattice decline survives composition matching, but unequal seat
counts imply slightly different finite-sample plug-in bias; the
comparison is exploratory and mechanism-free. The high-temperature
continuation interaction was not registered, and the Gemini tier is
descriptive under endpoint non-stationarity. Human-comparison limits are
detailed in §4.1. Post-adjudication family analyses cannot create
retrospective confirmation; §4.4 gives the complete power-bounded p13
status.

\subsection{7. Reproducibility and data
availability}\label{reproducibility-and-data-availability}

The public repository contains the event stores, prompt registries,
sealed registrations, adjudication records, timestamp proofs,
post-adjudication analyses, figures, manuscript history, and review
record. The one-command capsule verifies all 4,916 confirmatory Phase
3--5 runs with zero credentials and zero live model calls. The audit
covers 320 registered Phase 3/X1 LLM runs, 20 deterministic Phase 3
baselines, 2,864 Phase 4 runs, and 1,712 Phase 5 runs; three additional
completed legacy entry/diagnostic runs are also replayed but are not
counted as confirmatory. Phase 3 replay re-renders every prompt,
requires recorded-cache hash hits, reparses raw completions, recomputes
actions, payoffs, and RNG draw counts, and checks recorded call parity.
The deterministic P3-C3 baseline is independently recomputed from
archived seeds and game objects.

Phase 3 used a legacy provider path without Phase 4--5 response IDs or
deterministic request-body SHA capture. Phase 4--5 contain 30,421 normal
request events and 30,397 response events; the 24-event difference is
the disclosed provider-failure partial set. Each partial belongs to a
failed run, was excluded from completed-run analyses and the 4,916-run
replay denominator, and was never decoded as an action. Request attempts
remain represented in request and budget accounting rather than being
silently discarded; only completed replacement runs, where present under
the registered replacement procedure, enter analysis. Those response
records contain rendered prompts, bundle and request-body hashes, engine
commit and provider route, raw text, and provider response IDs.
Individual completion payloads were not provider-attested or separately
hash-chained at receipt. Capsule checksum manifests and external
timestamps make the released database snapshot tamper-evident relative
to publication; replay cannot prove that no alteration occurred before
snapshot sealing.

Reproduce the confirmatory record with:

\begin{Shaded}
\begin{Highlighting}[]
\FunctionTok{git}\NormalTok{ clone https://github.com/yoheinakajima/synthetic{-}players}
\BuiltInTok{cd}\NormalTok{ synthetic{-}players/capsule}
\FunctionTok{bash}\NormalTok{ verify.sh}
\end{Highlighting}
\end{Shaded}

All post-adjudication analyses are zero-call scripts over the archived
databases and are labeled separately from prospectively registered
results. The principal machine-readable surfaces are the Submission
Analyses directory (count reconciliation, dependence sensitivities,
family audits, and the final prior sweep), the Persona Table, the
Literature Map and Novelty Relationships documents, and the Review
Archive. Stable links to each are provided from the repository's review
entry point, keeping the empirical argument readable without embedding
long filesystem paths in the paper.

\subsection{8. Attribution}\label{attribution}

The human author selected the research questions, approved the
registered designs, adjudicated which reviewer recommendations to adopt,
and accepts responsibility for every claim. The autonomous pipeline
executed registration, dispatch, adjudication, and replay as apparatus.
AI reviewers supplied adversarial analysis that materially changed the
manuscript, including the family-error diagnosis, the condition-mean
identity, the dependence-unit critique, and the latent-person-invariance
correction.

The role of one round-2 reviewer subsequently expanded from critique to
specification of post-adjudication analyses and management of their
integration; a later reviewer independently checked the resulting branch
through an anonymous blobless clone against the generated JSON and
commit history. The analyses were executed by GitHub Actions against
archived databases, with commits authored by Yohei Nakajima or the
Actions bot. AI systems are not listed as authors. Their roles as
research apparatus and adversarial reviewers are disclosed here and in
the public review record (§7); the human author accepts responsibility
for every claim.

\begin{center}\rule{0.5\linewidth}{0.5pt}\end{center}

\subsection{Appendix A --- Supplementary scope and prospective
design}\label{appendix-a-supplementary-scope-and-prospective-design}

\subsubsection{A.1 Temperature
secondary}\label{a.1-temperature-secondary}

Choice entropy is defined as base-2 Shannon entropy,
\(H=-\sum_a p(a)\log_2p(a)\), over round-one payoff-role choices. The
historical registered secondary pooled all valid choices at each
temperature, but the T=0.7 and higher-temperature samples had different
composition. The matched-sweep reanalysis uses only persona-cell lanes
observed at all three temperatures:

\begin{longtable}[]{@{}
  >{\raggedleft\arraybackslash}p{(\columnwidth - 8\tabcolsep) * \real{0.2000}}
  >{\raggedleft\arraybackslash}p{(\columnwidth - 8\tabcolsep) * \real{0.2000}}
  >{\raggedleft\arraybackslash}p{(\columnwidth - 8\tabcolsep) * \real{0.2000}}
  >{\raggedleft\arraybackslash}p{(\columnwidth - 8\tabcolsep) * \real{0.2000}}
  >{\raggedleft\arraybackslash}p{(\columnwidth - 8\tabcolsep) * \real{0.2000}}@{}}
\toprule\noalign{}
\begin{minipage}[b]{\linewidth}\raggedleft
temperature
\end{minipage} & \begin{minipage}[b]{\linewidth}\raggedleft
matched units
\end{minipage} & \begin{minipage}[b]{\linewidth}\raggedleft
seats
\end{minipage} & \begin{minipage}[b]{\linewidth}\raggedleft
pooled Shannon entropy (bits)
\end{minipage} & \begin{minipage}[b]{\linewidth}\raggedleft
mean within-unit entropy (bits)
\end{minipage} \\
\midrule\noalign{}
\endhead
\bottomrule\noalign{}
\endlastfoot
0.7 & 13 & 544 & 0.8310 & 0.4484 \\
1.0 & 13 & 284 & 0.7822 & 0.2566 \\
1.3 & 13 & 284 & 0.7698 & 0.2877 \\
\end{longtable}

The registered pooled decline is partly composition-confounded but
survives on the identical sweep lattice. Mean within-unit entropy is
reported separately because pooled entropy can remain high when
different prompt-cell units occupy opposite boundaries. Neither
statistic identifies a temperature mechanism. The plug-in Shannon
estimates use unequal seat counts (544 versus 284 and 284), so their
small finite-sample biases are not identical; no bias-corrected entropy
claim is made.

\subsubsection{A.2 Other supplementary
findings}\label{a.2-other-supplementary-findings}

The following registered or explicitly descriptive results are outside
the paper's main composition--response arc but are retained here so that
the completed experimental record is citable.

\paragraph{Opponent-contingent adversarial play.}
In the 50-round RPS adversary suite, a second-order n-gram opponent earned
+0.215 payoff units per round against GPT-4.1 and survived Holm correction.
A first-order tracker instead earned -0.118, meaning the subject beat it.
The WSLS-targeter, designed around the earlier near-deterministic
lose-shift signature, earned only +0.008 with a lower confidence bound
below zero, although it outperformed the first-order tracker by +0.126.
The shuffled-history control was not worse than the ordered tracker. Thus
a behavioral signature measured against one opponent did not transport as
a simple profitable rule against another; exploitable structure was
opponent-contingent rather than captured by a single global label. The
GPT-4.1 opponent tests were registered; the six secondary arm tests were
Holm-controlled, while the cross-opponent transport interpretation here is
descriptive.

\paragraph{Role-attached asymmetry in a symmetric game.}
After RPS moves were renamed with neutral symbols and display order was
exactly counterbalanced, the registered GPT-4.1 first-minus-rock contrast
was -0.181: the preregistered positive-direction hypothesis was not
supported, and the observed direction was reversed. A support-only
Dirichlet analysis assigned probability 0.0001 to first-only selection
exceeding rock-only selection. The cross-vendor mirror had the opposite
sign, +0.243, with a one-sided lower bound of +0.139. The GPT-4.1 contrast
was registered; the Gemini mirror is reported descriptively under the
paper's cross-vendor tier. The pattern is descriptive evidence of
vendor-specific, role-attached asymmetry in a formally symmetric game,
not a universal ``rock bias.''

\paragraph{Cross-vendor label--payoff dissociation.}
In the canonical label-swap cell, GPT-4.1 chose the cooperation role 1.000
of the time while choosing the displayed word COOPERATE 0.000, thereby
following the displayed token DEFECT even when it named the
payoff-dominated role. Gemini chose the cooperation role 0.213 and the
word COOPERATE 0.788, mostly following the payoff-dominant role instead.
When counterfactual payoffs made the cooperation role strictly dominant
and displayed it as DEFECT, the two models' cooperation-role shares were
1.000 and 0.975. For GPT-4.1, word and payoff dominance were congruent in
this cell, so it does not separate the two channels. For Gemini, the cell
is informative because the model moved to the payoff-dominant cooperation
role despite that role carrying DEFECT. The GPT-4.1 cells were registered;
the Gemini figures are descriptive under the paper's cross-vendor tier.
The vendors differed in when a familiar action word overrode the supplied
payoff structure; the data do not identify one universal lexical
mechanism.

\paragraph{Endpoint drift and subject eligibility.}
The Gemini sentinel fingerprint changed from 10/10 baseline matches to a
sequence reaching 6/10 and 7/10 on the unversioned endpoint, triggering
freezes, disclosure, re-baselining, and a later attestation gate. GPT-4.1's
sentinel cells remained 10/10. Separately, Claude Haiku failed the
registered behavioral entry gate and was replaced through a sealed
amendment. This is a procedural record, not a behavioral-effect estimate.
Together the cases show that model identity and endpoint availability do
not by themselves establish a stable, usable behavioral subject.

\subsubsection{A.3 Prospective
replication}\label{a.3-prospective-replication}

A Phase 6 replication will preselect one target or a small candidate
family and preregister the complete familywise procedure: candidate set,
episode-level unit, interiority gate, maximum statistic, decision
threshold, and sample size. For the composition estimand, it will either
widen the preregistered prompt family or prospectively sample from a
declared persona generator rather than reuse the fixed sixteen prompts
as if they represented a population. For the response estimand, the
episode allocation will be chosen by simulating the exact registered
gate and candidate family until the minimum attainable familywise
p-value is below the declared threshold and power meets a prespecified
target under the declared alternative. The design should also include a
format-matched neutral prefix, a continuation-probability × wording
factorial, and a structurally equivalent non-PD label-conflict control.

\subsubsection{A.4 Research record}\label{a.4-research-record}

The complete correction ledger, sealed discussion text, dead-predictions
ledger, reviewer-role disclosures, and mechanical disposition matrices
are maintained in the public Review Archive and Submission Analyses
directories. Persona provenance is documented in the Persona Table;
scholarly positioning is documented in the Literature Map and Novelty
Relationships documents. Historical artifacts are never silently
rewritten; current interpretations are linked to the versions they amend.

\subsection{References}\label{references}

\begingroup
\small
\setlength{\parskip}{0.30em}

Berger, J. O., Bernardo, J. M., and Sun, D. (2009). The formal
definition of reference priors. \emph{Annals of Statistics, 37}(2),
905--938. https://doi.org/10.1214/07-AOS587

Efron, B., and Tibshirani, R. J. (1993). \emph{An Introduction to the
Bootstrap}. Chapman \& Hall/CRC.
https://doi.org/10.1007/978-1-4899-4541-9

Holtzman, A., Buys, J., Du, L., Forbes, M., and Choi, Y. (2020). The
curious case of neural text degeneration. In \emph{International
Conference on Learning Representations}.

Clopper, C. J., and Pearson, E. S. (1934). The use of confidence or
fiducial limits illustrated in the case of the binomial.
\emph{Biometrika, 26}(4), 404--413.
https://doi.org/10.1093/biomet/26.4.404

Hoeffding, W. (1940). Maßstabinvariante Korrelationstheorie.
\emph{Schriften des Mathematischen Instituts und des Instituts für
Angewandte Mathematik der Universität Berlin, 5}, 181--233.

Lehmann, E. L., and Romano, J. P. (2005). \emph{Testing Statistical
Hypotheses} (3rd ed.). Springer. https://doi.org/10.1007/0-387-27605-X

Sklar, A. (1959). Fonctions de répartition à n dimensions et leurs
marges. \emph{Publications de l'Institut de Statistique de l'Université
de Paris, 8}, 229--231.

Westfall, P. H., and Young, S. S. (1993). \emph{Resampling-Based
Multiple Testing: Examples and Methods for p-Value Adjustment}. Wiley.

Akata, E., Schulz, L., Coda-Forno, J., Oh, S. J., Bethge, M., and
Schulz, E. (2025). Playing repeated games with large language models.
\emph{Nature Human Behaviour, 9}, 1380--1390.
https://doi.org/10.1038/s41562-025-02172-y

Anthis, J. R., Liu, R., Richardson, S. M., Kozlowski, A. C., Koch, B.,
Brynjolfsson, E., Evans, J., and Bernstein, M. S. (2025). Position: LLM
social simulations are a promising research method. \emph{Proceedings of
the 42nd International Conference on Machine Learning, PMLR 267},
81005--81034. https://proceedings.mlr.press/v267/anthis25a.html

Argyle, L. P., Busby, E. C., Fulda, N., Gubler, J. R., Rytting, C., and
Wingate, D. (2023). Out of one, many: Using language models to simulate
human samples. \emph{Political Analysis, 31}(3), 337--351.
https://doi.org/10.1017/pan.2023.2

Ashokkumar, A., Hewitt, L., Ghezae, I., and Willer, R. (2026). Large
language models can predict the results of social science experiments.
\emph{Nature}. https://doi.org/10.1038/s41586-026-10742-x

Batzner, J., Stocker, V., Tang, B., Natarajan, A., Chen, Q., Schmid, S.,
and Kasneci, G. (2025). Whose personae? Synthetic persona experiments in
LLM research and pathways to transparency. \emph{Proceedings of the
AAAI/ACM Conference on AI, Ethics, and Society, 8}(1), 343--354.
https://doi.org/10.1609/aies.v8i1.36553

Bisbee, J., Clinton, J. D., Dorff, C., Kenkel, B., and Larson, J. M.
(2024). Synthetic replacements for human survey data? The perils of
large language models. \emph{Political Analysis, 32}(4), 401--416.
https://doi.org/10.1017/pan.2024.5

Boelaert, J., Coavoux, S., Ollion, É., Petev, I., and Präg, P. (2025).
Machine bias: How do generative language models answer opinion polls?
\emph{Sociological Methods \& Research, 54}(3), 1156--1196.
https://doi.org/10.1177/00491241251330582

Cronbach, L. J., and Meehl, P. E. (1955). Construct validity in
psychological tests. \emph{Psychological Bulletin, 52}(4), 281--302.
https://doi.org/10.1037/h0040957

Dal Bó, P., and Fréchette, G. R. (2011). The evolution of cooperation in
infinitely repeated games: Experimental evidence. \emph{American
Economic Review, 101}(1), 411--429.
https://doi.org/10.1257/aer.101.1.411

Georgousis, D., Lymperaiou, M., Dimitriou, A., Filandrianos, G., and
Stamou, G. (2026). Evaluating counterfactual strategic reasoning in
large language models. arXiv:2603.19167.
https://doi.org/10.48550/arXiv.2603.19167

Grimm, V., Revilla, E., Berger, U., et al.~(2005). Pattern-oriented
modeling of agent-based complex systems: Lessons from ecology.
\emph{Science, 310}(5750), 987--991.
https://doi.org/10.1126/science.1116681

Harry, T., Ngong, I. C., Nweke, C., Feng, Y., and Near, J. (2026).
Beyond fixed psychological personas: State beats trait, but language
models are state-blind. In \emph{Findings of the Association for
Computational Linguistics: ACL 2026}, 26440--26468. Association for
Computational Linguistics.
https://doi.org/10.18653/v1/2026.findings-acl.1316

Horton, J. J. (2023). Large language models as simulated economic
agents: What can we learn from Homo Silicus? NBER Working Paper 31122.
https://doi.org/10.3386/w31122

Hullman, J., Broska, D., Sun, H., and Shaw, A. (2026). This human study
did not involve human subjects: Validating LLM simulations as behavioral
evidence. arXiv:2602.15785. https://doi.org/10.48550/arXiv.2602.15785

International Council for Harmonisation. (2000). \emph{ICH E10: Choice
of control group and related issues in clinical trials}.
https://database.ich.org/sites/default/files/E10\_Guideline.pdf

Li, Y., and Ji, X. (2026). When simulations look right but causal
effects go wrong: Large language models as behavioral simulators.
arXiv:2604.02458. https://doi.org/10.48550/arXiv.2604.02458

Lin, V., Yun, T., Matarić, M. J., Canny, J., Gretton, A., and D'Amour,
A. (2026). The illusion of intervention: Your LLM-simulated experiment
is an observational study. arXiv:2605.20767.
https://doi.org/10.48550/arXiv.2605.20767

Lucas, R. E., Jr.~(1976). Econometric policy evaluation: A critique.
\emph{Carnegie-Rochester Conference Series on Public Policy, 1}, 19--46.
https://doi.org/10.1016/S0167-2231(76)80003-6

Mei, Q., Xie, Y., Yuan, W., and Jackson, M. O. (2024). A Turing test of
whether AI chatbots are behaviorally similar to humans.
\emph{Proceedings of the National Academy of Sciences, 121}(9),
e2313925121. https://doi.org/10.1073/pnas.2313925121

Mousavi Davoudi, S. P., Amiri-Margavi, A., Gholami Davodi, A., Hasani
Balyani, H., and Gharagozlou, A. (2026). Same game, different story: A
minimal conservative strategic robustness benchmark for large language
model agents. arXiv:2607.19670.
https://doi.org/10.48550/arXiv.2607.19670

Pal, S., Mallela, A., Hilbe, C., Pracher, L., Wei, C., Fu, F., Schnell,
S., and Nowak, M. A. (2026). Strategies of cooperation and defection in
five large language models. arXiv:2601.09849.
https://doi.org/10.48550/arXiv.2601.09849

Park, J. S., Zou, C. Q., Kamphorst, J., Egan, N., Shaw, A., Hill, B. M.,
Cai, C., Morris, M. R., Liang, P., Willer, R., and Bernstein, M. S.
(2024, revised 2026). LLM agents grounded in self-reports enable
general-purpose simulation of individuals. arXiv:2411.10109v3.
https://doi.org/10.48550/arXiv.2411.10109

Persson, E., Schultzberg, M., and Ankargren, S. (2026). Statistical
foundations of LLM-based A/B testing: A surrogacy framework for human
causal inference. arXiv:2606.17165.
https://doi.org/10.48550/arXiv.2606.17165

Sclar, M., Choi, Y., Tsvetkov, Y., and Suhr, A. (2024). Quantifying
language models' sensitivity to spurious features in prompt design, or:
How I learned to start worrying about prompt formatting. In
\emph{International Conference on Learning Representations}.

Shanahan, M., McDonell, K., and Reynolds, L. (2023). Role play with
large language models. \emph{Nature, 623}, 493--498.
https://doi.org/10.1038/s41586-023-06647-8

Temple, R., and Ellenberg, S. S. (2000). Placebo-controlled trials and
active-control trials in the evaluation of new treatments. Part 1:
Ethical and scientific issues. \emph{Annals of Internal Medicine,
133}(6), 455--463.
https://doi.org/10.7326/0003-4819-133-6-200009190-00014

Windrum, P., Fagiolo, G., and Moneta, A. (2007). Empirical validation of
agent-based models: Alternatives and prospects. \emph{Journal of
Artificial Societies and Social Simulation, 10}(2), 8.
https://www.jasss.org/10/2/8.html

Xiao, Y., Zhang, V. J., Yang, C., Ma, N., Xuan, W., and Huang, J.-t.
(2026). The chameleon's limit: Investigating persona collapse and
homogenization in large language models. arXiv:2604.24698.
https://doi.org/10.48550/arXiv.2604.24698

Xie, Y., Liang, L., Li, S., Lu, Y., Xiao, Z., Shi, M., Huang, J., Wang,
M., and Xie, Y. (2026). Evaluating the statistical realism of
LLM-generated social science data. \emph{Proceedings of the National
Academy of Sciences, 123}(19), e2538145123.
https://doi.org/10.1073/pnas.2538145123

\endgroup

\end{document}